\documentclass[11pt]{article}
\usepackage[final]{acl}

\usepackage{times}
\usepackage{latexsym}
\usepackage{CJKutf8}
\usepackage{array}
\usepackage[T1]{fontenc}
\usepackage{microtype}
\usepackage{inconsolata}
\usepackage{algorithm}
\usepackage{algorithmic}
\usepackage{graphicx}
\usepackage{multirow}
\usepackage{amsmath}

\title{Efficiently Exploring Large Language Models for Document-Level Machine Translation with In-context Learning}

\author{
Menglong Cui{\thanks{~Equal contribution}}, Jiangcun Du{\footnotemark[1]}, Shaolin Zhu, Deyi Xiong {\thanks{~Corresponding author}} \\
  College of Intelligence and Computing, Tianjin University, Tianjin, China\\
  \texttt{\{cuimenglongcs,d2000,zhushaolin,dyxiong\}@tju.edu.cn} \\}

\begin{document}
\begin{CJK}{UTF8}{gbsn}

\maketitle

\begin{abstract}

Large language models (LLMs) exhibit outstanding performance in machine translation via in-context learning.
In contrast to sentence-level translation, document-level translation (DOCMT) by LLMs based on in-context learning faces two major challenges: (1) document translations generated by LLMs are often incoherent; (2) the length of demonstration for in-context learning is usually limited.
To address these issues, we propose a \textbf{C}ontext-\textbf{A}ware \textbf{P}rompting method (\textbf{CAP}), which enables LLMs to generate more accurate, cohesive, and coherent translations via in-context learning.
CAP takes into account multi-level attention, selects the most relevant sentences to the current one as context, and then generates a summary from these collected sentences. Subsequently, sentences most similar to the summary are retrieved from the datastore as demonstrations, which effectively guide LLMs in generating cohesive and coherent translations.
We conduct extensive experiments across various DOCMT tasks, and the results demonstrate the effectiveness of our approach, particularly in zero pronoun translation (ZPT) and literary translation tasks.

\end{abstract}

\section{Introduction}

With the increasing scale of parameters and training corpus, large language models (LLMs) have gained remarkable abilities to handle a variety of tasks via in-context learning \cite{DBLP:journals/corr/abs-2310-09748,DBLP:journals/corr/abs-2307-01137}, which allows language models to perform tasks with a few given demonstrations and human-written prompts as context. 
One particular area where LLMs have shown outstanding potential is machine translation (MT) \cite{DBLP:conf/eamt/MoslemHKW23,DBLP:conf/icml/0006HB23,DBLP:journals/corr/abs-2306-10968,DBLP:journals/corr/abs-2401-05861}. 
However, most existing studies \cite{DBLP:conf/coling/ZhuCX24} have focused on sentence-level translation, which makes LLM-based document-level machine translation underexplored \cite{DBLP:conf/emnlp/WangLJZY0T23,DBLP:journals/corr/abs-2401-08088}.
To bridge this gap, we adapt LLMs to document-level machine translation (DOCMT) in this paper.

As a complex task different from sentence-level translation, one major challenge of DOCMT with LLMs is that \textbf{the length of demonstrations for in-context learning is limited}.
For sentence-level MT, \citet{DBLP:conf/acl/AgrawalZLZG23} and \citet{DBLP:conf/icml/0006HB23} show that using 32 or more randomly sampled bilingual parallel sentence pairs as prompts can effectively enhance the translation abilities of LLMs.
However, for DOCMT, the length of the text segments to be translated or to be used as demonstrations inherently increase \cite{DBLP:journals/corr/abs-2401-03868}. 
Consequently, the limited input window of large language models poses a significant obstacle \cite{DBLP:journals/corr/abs-2401-07872,DBLP:conf/acl/TangKH023}. 
Unlike sentence-level translation, the constraints of DOCMT hinder the seamless integration of such demonstrations.
Moreover, the lengthier nature of document-level inputs exacerbates issues related to inference speed. 
As the length of the demonstrations increases, the computational resources required for processing also escalate, potentially impeding real-time translation capabilities.

Previous efforts to DOCMT with LLMs, e.g., the method presented in \cite{DBLP:journals/corr/abs-2401-06468}, involve fine-tuning LLMs using two strategies: Parameter-Efficient Fine-Tuning and Full Parameter Fine-Tuning.
Such a framework may not be able to sufficiently overcome the limitation of the length of demonstration for in-context learning in LLMs.
% off-target
And it also reveals a significant issue: \textbf{context-independent translations}, which can result in translations that lack coherence and disambiguation \cite{DBLP:conf/acl/ZhangBJDAF22,DBLP:conf/iwslt/MaceS19}.

To address these issues, we propose a novel method called \textbf{CAP}, which stands for \textbf{C}ontext-\textbf{A}ware \textbf{P}rompting. 
It consists of three essential steps. Firstly, by combining token- and sentence-level attention scores, we obtain sentences with the highest sentence-level attention scores as context for each sentence. This step facilitates LLMs in generating coherent translations. Secondly, to address the limitation of the length of demonstration for in-context learning in LLMs, we summarize the context and retrieve parallel sentences from the datastore, which are most similar to the generated summary. Finally, we use the retrieved example pairs as demonstrations, prompting LLMs to produce document translations.

In summary, the contributions of this work are as follows:
\begin{itemize}

\item 
We investigate document-level translation via LLMs. Specially, we explore linguistic phenomena in document-level translation with LLMs, such as ZPT.
\item We propose a novel method, CAP, to enable LLMs to perform accurate and coherent translation via in-context learning. Our method addresses the issue of translation coherence deficiency by employing a dynamic context window and resolves the limitation concerning the length of demonstration for in-context learning by summarizing the context.
\item To demonstrate the effectiveness of our method, we conduct experiments on various DOCMT tasks. Experimental results show that our method significantly outperforms the baselines, especially for the ZPT task.
\end{itemize}

\section{Related Work}

\textbf{Document-Level Machine Translation}
In recent years, numerous approaches have been proposed for DOCMT \cite{DBLP:journals/csur/MarufSH21,DBLP:conf/coling/LeiRX22,DBLP:conf/icassp/ZhangZCLS22,DBLP:conf/ijcai/TanZ0Z22,DBLP:conf/ecai/ZhangZCYWX20}.
These studies could be further roughly categorized into two groups. 
Studies of the first group extend the source-side context from a sentence to the local context of few sentences \cite{DBLP:conf/emnlp/TanZZ21,DBLP:conf/emnlp/LyuLGZ21,DBLP:conf/emnlp/XuLW0C21}.
Studies of the second group have focused on document-to-document (Doc2Doc) translation. 
The initial exploration of Doc2Doc NMT has involved concatenating multiple sentences into a single unit \cite{DBLP:conf/acl/MaZZ20,DBLP:conf/emnlp/ZhangCGF20,DBLP:conf/emnlp/TanZXZ19}.
Recent studies have successfully trained vanilla Transformer models for Doc2Doc translation by leveraging either large augmented datasets \cite{DBLP:conf/wmt/LupoDB22,DBLP:conf/acl/SunWZZHCL22} or pre-trained models \cite{DBLP:conf/naacl/YangZCXL21}. 
Additionally, some latest work has achieved the truly Doc2Doc translation by enhancing the attention mechanism in the Transformer model \cite{DBLP:conf/acl/Bao0TCL20,DBLP:journals/taslp/LiLJTYZ23}.

\textbf{Translation Oriented LLMs} LLMs have demonstrated remarkable proficiency
across a wide range of NLP tasks \cite{DBLP:journals/corr/abs-2211-05100,DBLP:journals/corr/abs-2302-13971,DBLP:conf/coling/SunX22,DBLP:journals/csur/MinRSVNSAHR24,DBLP:journals/tmlr/WeiTBRZBYBZMCHVLDF22}.
Recent research has shown that in-context learning can significantly enhance their performance when following general language instructions.
Specifically, there is a growing body of work exploring the translation capabilities of LLMs \cite{DBLP:journals/corr/abs-2305-06575,DBLP:conf/icml/0006HB23,DBLP:journals/corr/abs-2302-09210}. 
However, it is important to note that these efforts have primarily focused on sentence-level machine translation and have not delved into DOCMT \cite{DBLP:journals/corr/abs-2401-06468}.
Latest noteworthy studies in DOCMT are conducted by \citet{DBLP:journals/corr/abs-2401-06468} and \citet{DBLP:conf/emnlp/WangLJZY0T23}.
\citet{DBLP:conf/emnlp/WangLJZY0T23} investigate the document-level translation capabilities of GPT-3.5-TURBO, making it the most closely related work to ours that explores the DOCMT ability of LLMs.
\citet{DBLP:journals/corr/abs-2401-06468} explore fine-tuning LLMs using two strategies: Parameter-Efficient Fine-Tuning and Full Parameter Fine-Tuning, to enhance the DOCMT ability of LLMs.
Such a framework may not be able to sufficiently overcome the limitation of the length of demonstration for in-context learning in LLMs.
We observe that the existing translation-oriented LLMs tend to struggle with effectively translating sentences in the middle and ending of a document.
Existing methods that explore DOCMT with LLMs are rarely considering the issues of (1) the length limitation of demonstrations for in-context learning and (2) maintaining coherence and resolving ambiguities.
In this work, we mainly concentrate on mitigating the length limitation of in-context learning for DOCMT and addressing the issues of incoherence and ambiguity in DOCMT with LLMs.

\begin{figure*}[ht]
\begin{center}
\includegraphics[scale=0.8]{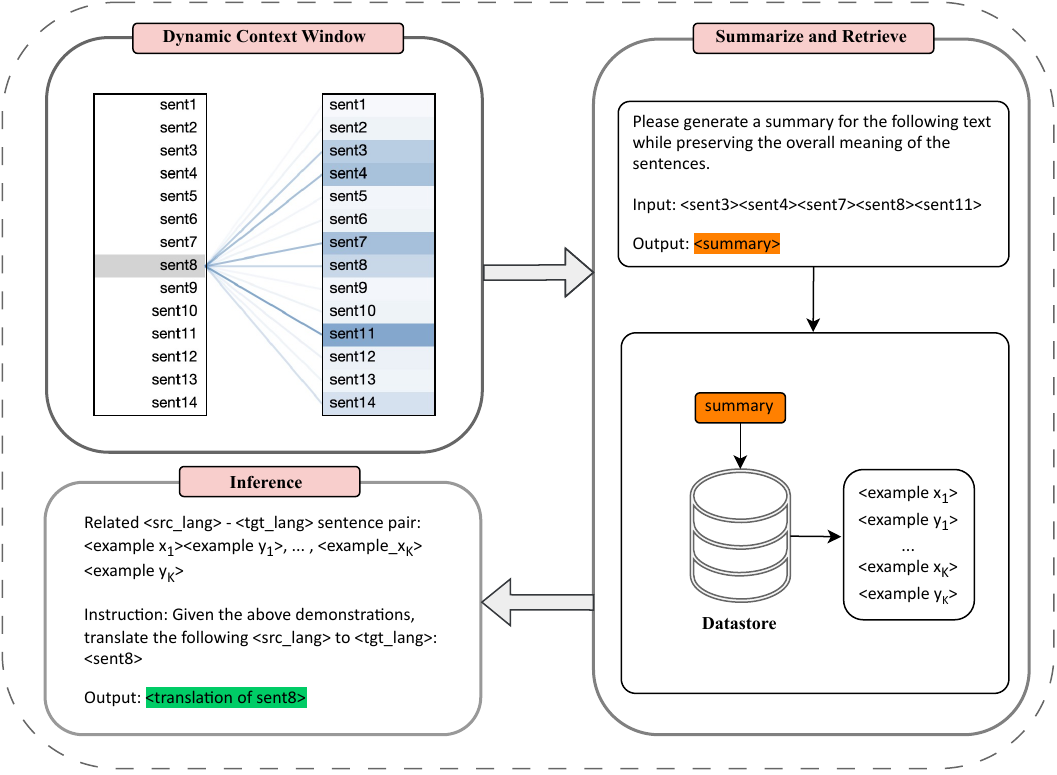} 
% (1)使用sentence-level attention来挑选context；(2) 
\caption{Diagram of the proposed CAP. It consists of three essential steps. (1) Dynamic Context Window: selecting context by sentence-level attention scores; (2) Summarize and Retrieve: summarizing the context and retrieving parallel sentences from the datastore, which are most similar to the summary; (3) Inference: with parallel sentences serving as demonstrations, the LLM translates the input text  via in-context learning.}
\label{ours}
\end{center}
\end{figure*}

% \section{Proposed Method}
\section{CAP: Context-Aware Prompting}
\label{Proposed Method}

In this section, to address the challenges associated with utilizing LLMs for DOCMT via in-context learning, we propose a \textbf{C}ontext-\textbf{A}ware \textbf{P}rompting method.
As shown in Figure \ref{ours}, our approach consists of three steps. 
Firstly, we take into consideration multi-level attention, subsequently opting for the sentences with the highest sentence-level attention scores as the context.
Secondly, we input the context into an LLM to obtain its summary, then retrieve $K$ example pairs from the datastore, which are most similar to the summary. 
Thirdly, we use the retrieved example pairs as demonstrations in the prompt, guiding the model to generate accurate and coherent translations through a few-shot approach. 

\subsection{Dynamic Context Window}
\label{Dynamic Context Window}

In the context of document-level translation for LLMs, issues of incoherence and disambiguation may arise, leading to context-independent translations. In order to address this problem, we propose a Dynamic Context Window as a solution.
By extracting the most attended context surrounding the current sentence, it facilitates the model in generating accurate and coherent translation outputs.

Initially, we meticulously devise a method for computing sentence-level attention. Specifically, as demonstrated in Eq.\;(\ref{eq1}), we extract the attention weights from the final layer of the LLM, followed by averaging across all attention heads to obtain the token-token attention scores.
Next, as illustrated in Eq.\;(\ref{token_sent_socre}), the maximum value of token-token attention scores between tokens in the current sentence and tokens in all other sentences is computed as the token-sentence attention score. 
Ultimately, as shown in Eq.\;(\ref{sent_sent_socre}), the average of attention scores between all tokens in the current sentence and other sentences is calculated to derive the sentence-level attention score.

\begin{equation}
\boldsymbol{attn}_{i, j} = \frac{1}{H}\sum\limits_{h=1}^{H} \boldsymbol{attn}^{h}_{i,j} 
\label{eq1}
\end{equation}
\begin{equation}
\rm{TS_{i, S}}=max(\boldsymbol{attn}_{i, k}), k \in tokens(S) \label{token_sent_socre}
\end{equation}
\begin{equation}
\rm{SS_{I, S}}=avg(TS_{i, S}), i \in tokens(I) \label{sent_sent_socre}
\end{equation}

\noindent where $\rm{TS}_{i, S}$ is the attention score of \textbf{T}oken $\rm{i}$ and \textbf{S}entence $\rm{S}$, $\rm{SS}_{I, S}$ is the attention score between \textbf{S}entence $\rm{I}$ and \textbf{S}entence $\rm{S}$. 
$\boldsymbol{attn}_{i, j}^{h}$ denotes the attention score of the $h$-th head in the final transformer block, calculated between the $i$-th token and the $j$-th token in the current document.
$H$ is the number of attention heads in each transformer block.

After obtaining the sentence-level attention scores, we extract the top $N$ sentences with the highest sentence-level attention scores as dynamic context, while maintaining their original sequential positions within the document.

\subsection{Example-Specific In-Context Demonstrations Selection}
\label{Example-specific In-context Examples}

Unlike sentence-level translation, the challenge in utilizing LLMs for DOCMT arises from the constrained context length, as the length of the text segments to be translated or used as demonstrations inherently increases.
% limiting in-context learning and impeding nuanced capture of document context. 
To effectively address these challenges, it is necessary to further process the context and extract as much relevant information as possible from the preceding sentences. We then input this extracted context into an LLM to generate a concise summary that encapsulates key topics and information.

Subsequently, as shown in Eq.\;(\ref{eq4}), we utilize sentence-transformers \cite{DBLP:journals/corr/abs-2004-09813} to retrieve carefully designed example pairs from a specified datastore. These examples are selected based on their cosine similarity with the embedding of the summary sentences, ensuring that the retrieved examples exhibit semantic similarity to the context and content of the document.
\begin{equation}
{\rm {score}}(S, X) = cos(\boldsymbol{S}, \boldsymbol{X}) \label{eq4}
\end{equation}
\noindent where $S$, and $X$ are the summary we obtained at the preceding step and the example sentence in datastore respectively. $\boldsymbol{S}$ and $\boldsymbol{X}$ are the embeddings of $S$ and $X$ respectively. 

By employing these selected examples as few-shot demonstrations, we provide the LLM with a more precise understanding of the document. This approach not only enhances the model's accuracy in utilizing in-context learning for translation but also contributes to enhancing the robustness of document translation tasks.

\subsection{Inference}
\label{Inferring and evaluating}

We employ a prompt template, which is identical to that utilized in \cite{DBLP:journals/corr/abs-2305-04118}. In this approach, we integrate the retrieved example pairs from the preceding step seamlessly into the prompt template, which will then be input into the target LLM. This strategic integration serves to guide the LLMs in generating translations that exhibit enhanced cohesion and coherence.

Following the translation process, we employ a diverse set of evaluation metrics to thoroughly assess the quality of the translations generated by the LLM. This multifaceted evaluation allows us to gain comprehensive insights into the effectiveness and performance of the language models in producing accurate and appropriate translations.

\section{Experiments}
In this section, we conducted extensive experiments on various DOCMT tasks to examine the effectiveness of the proposed method.

\subsection{Setup}

\textbf{Datasets:} We used OPUS-100\footnote{https://opus.nlpl.eu/opus-100.php} and news-commentary-v15 as the datastore to retrieve the in-context demonstrations on four language pairs. 
To assess the effectiveness of our proposed method on LLMs, we employed WMT22 newstest as the test set.
Following \cite{DBLP:conf/acl/AgrawalZLZG23}, we normalized punctuation using Moses\footnote{http://www2.statmt.org/moses/} and removed sentence pairs with a source/target length ratio exceeding 1.5.
To assess the accuracy of our method in ZPT, we utilized the GuoFeng \cite{DBLP:conf/emnlp/XuWWLSC0T22} dataset, which covers five domains: movie subtitle, Q\&A forum, government news, web fiction, and personal profile.
To validate the capability of our approach in guiding the model to generate more cohesive and more coherent translations, we employed the WMT23 Literary Translation dataset,\footnote{https://www2.statmt.org/wmt23/literary-translation-task.html} which comprises two test sets: an in-domain test set and an out-of-domain test set. 
Additionally, we used its training set to retrieve in-context demonstrations.

\begin{table*}[ht]
\centering
\begin{tabular}{lcccccccc}
\hline
\multicolumn{1}{l|}{\multirow{2}{*}{methods}}   & \multicolumn{2}{c}{Qwen-1.8B-Chat} & \multicolumn{2}{c}{Qwen-7B-Chat} & \multicolumn{2}{c}{Qwen-14B-Chat} & \multicolumn{2}{c}{Qwen-72B-Chat}  \\
\cline{2-9}
\multicolumn{1}{l|}{} & de-en & zh-en & de-en & zh-en & de-en & zh-en & de-en & zh-en \\
 \hline
\multicolumn{1}{l|}{Zero-shot} & \textbf{19.84} & \textbf{17.28} & 34.21 & 24.02 & 35.99 & 26.37 & 37.42 & 29.11 \\
\multicolumn{1}{l|}{Random}    & 17.58 & 15.12 & 32.91 & 23.58 & 35.28 & 26.42 & 37.55 & 28.40 \\
\multicolumn{1}{l|}{BM25}    & 17.70 & 14.92 & 32.90 & 23.54 & 34.86 & 25.90 & 37.49 & 28.59 \\
\multicolumn{1}{l|}{Similar}   & 17.81 & 15.34 & 33.43 & 23.85 & 35.29 & 
26.42 & 37.72 & 28.50 \\
\multicolumn{1}{l|}{Precedent} & 15.80 & 15.23 &  34.49 & 23.65 & 35.75 & 26.49 & 38.01 & 29.15 \\
\multicolumn{1}{l|}{Ours}      & 18.56 & 15.51 & \textbf{34.86} & \textbf{24.34} & \textbf{36.57} & \textbf{26.77} & \textbf{38.14} & \textbf{29.32} \\
\hline
\end{tabular}
\caption{Document-level BLEU scores of variants of the Qwen chat version LLM, employing different prompt strategies, on the WMT22 newstest dataset.}
\label{tab-main-results-1}
\end{table*}

\textbf{Models:} To compare the translation performance of LLMs using our method, we compared against the following three encoder-decoder based strong NMT models as baselines.
\begin{itemize}
\item MarianMT: A framework for translation models, using the same models as BART \cite{DBLP:conf/acl/LewisLGGMLSZ20}. 
In our experiment, each language pair is associated with a distinct pre-trained model.
\item M2M100 \cite{DBLP:journals/jmlr/FanBSMEGBCWCGBL21}: It is a multilingual encoder-decoder model, which is pre-trained on 100 languages and is capable of translating 9,900 directions.
\item NLLB \cite{DBLP:journals/corr/abs-2207-04672}: It is a neural multilingual machine translation model, which supports translation between any pair of 200 languages. There are numerous variants of it, and the one utilized in our experiments is NLLB-3.3B.
\end{itemize}

To evaluate the translation abilities of LLMs, we utilized LLMs that are tuned with SFT, as they can follow instructions to generate translations.
Hence, we chose the following two LLMs, which support both Chinese and English and have undergone instruction tuning:

\begin{itemize}
\item Baichuan \cite{DBLP:journals/corr/abs-2309-10305}: It is trained on a high-quality corpus with 2.6 trillion tokens and has achieved the best performance in authoritative Chinese and English benchmarks of the same size. 
\item Qwen \cite{DBLP:journals/corr/abs-2309-16609}: It is pretrained on over 3 trillion tokens, including Chinese, English, multilingual texts, code, and mathematics, covering general and professional fields. 
\end{itemize}

\textbf{Prompt Selection Strategies:} To assess the effectiveness of the proposed method, we compared our method against five prompt selection strategies: 
\begin{itemize}
\item Zero-shot: Zero-shot is a powerful baseline. Previous studies \cite{Radford2019LanguageMA} show that LLMs have excellent zero-shot abilities in many NLP tasks, including reading comprehension, translation, and summarization.

\item Random: This strategy randomly selects $K$ sentence pairs from the datastore as demonstrations.

\item BM25: 
It calculates the similarity between translated sentences and sentences in the corpus with BM25, and then selects top $K$ sentences with the highest similarity as demonstrations.

\item Similar: This method uses sentence embedding similarity-based retrieval to select $K$ demonstrations \cite{DBLP:conf/eamt/MoslemHKW23}. In our work, we used sentence-transformers \cite{DBLP:journals/corr/abs-2004-09813} to implement this method.

\item Precedent: This approach guides large language models in generating translations by utilizing the preceding $K$ sentences of the current sentence within a document, along with those translations generated in previous steps as demonstrations.
\end{itemize}

For prompt selection experiments, we employed models from four distinct sizes of the Qwen LLMs. Meanwhile, for a trade-off between performance and cost, we set $K$ to 3.

\textbf{Metrics:} We employed a variety of metrics to assess the quality of document-level translation generated by LLMs, including document-level BLEU($d$-BLEU), and ChrF2 using SacreBLEU for document-level evaluation. 
For the ZPT task, we employed accuracy to evaluate the precision of zero pronoun translation. 
Additionally, we have also evaluated document-level translation quality with Blonde \cite{DBLP:conf/naacl/JiangLM0YHSCS022}, which is an automatic evaluation metric for document-level machine translation.
More experimental results can be found in Appendix \ref{Experimental results}.

All experiments were performed on 4 NVIDIA A100 GPUs.

\begin{table*}[ht]
\centering
\scalebox{0.95}{
\begin{tabular}{lcccccccc}
\hline
\multicolumn{1}{l|}{\multirow{2}{*}{Methods}}  & \multicolumn{2}{c}{de-en} & \multicolumn{2}{c}{zh-en} & \multicolumn{2}{c}{en-de} & \multicolumn{2}{c}{en-zh} \\
\cline{2-9}
\multicolumn{1}{l|}{} & $d$-BLEU & chrF2 & $d$-BLEU & chrF2 & $d$-BLEU & chrF2 & $d$-BLEU & chrF2 \\
\hline
 & \multicolumn{8}{c}{Encoder-Decoder} \\
\hline
\multicolumn{1}{l|}{MarianMT}   & 36.25 & 68.28 & 22.15 & 58.60 & 30.99 & 58.29 & 27.08 & 24.89 \\
\multicolumn{1}{l|}{M2M100-1.2B}   & 35.11 & 66.86 & 17.32 & 49.15 & 31.13 & 67.51 & 28.45 & 26.13 \\
\multicolumn{1}{l|}{NLLB-3.3B}     & 36.35 & 67.74 & 25.48 & 60.03 & \textbf{34.08} & \textbf{69.23} & 24.03 & 23.02 \\
\hline
 & \multicolumn{8}{c}{LLMs} \\
\hline
\multicolumn{1}{l|}{Baichuan-7B-Chat} & 33.59 & 68.32 & 25.06 & 60.70 & 20.87 & 63.08 & 33.34 & 33.57 \\
\multicolumn{1}{l|}{Baichuan-13B-Chat} & 33.62 & 68.83 & 26.02 & 64.91 & 22.32 & 64.83 & 36.30 & 33.94 \\
\multicolumn{1}{l|}{Qwen-7B-Chat}  & 34.86 & 68.84 & 24.34 & 62.57 & 24.01 & 64.72 & 36.59 & 33.67 \\
\multicolumn{1}{l|}{Qwen-14B-Chat} & 36.57 & 70.00 & 26.77 & 64.24 & 29.36 & 67.88 & 42.28 & 37.85 \\
\multicolumn{1}{l|}{Qwen-72B-Chat} & \textbf{38.14} & \textbf{70.44} & \textbf{29.32} & \textbf{65.43} & 31.48 & 68.11 & \textbf{46.80} & \textbf{40.90} \\
\hline
\end{tabular}
}

\caption{Document-level BLEU and chrF2 scores of models with different architectures on four language pairs from the WMT22 newstest dataset.}
\label{encoder-decoder}
\end{table*}

\begin{table}[ht]
\centering
\begin{tabular}{lcccc}
\hline
methods   & \multicolumn{1}{c}{1.8B} & \multicolumn{1}{c}{7B} & \multicolumn{1}{c}{14B} & \multicolumn{1}{c}{72B}  \\
\hline
Zero-shot & 29.29          & 36.00          & 38.30          & 36.53          \\
Random    & 27.81          & 38.59          & 37.55          & 38.13          \\
BM25      & 28.13          & 37.45          & 37.94          & 37.95         \\
Similar   & 29.14          & 37.49          & 37.93          & 37.39          \\
Precedent & 29.91          & 42.02          & 40.45          & 39.07    \\
Ours      & \textbf{32.58} & \textbf{42.13} & \textbf{41.87} & \textbf{41.65} \\
\hline
\end{tabular}
\caption{ZPT accuracy of various variants of the Qwen chat version model on the GuoFeng dataset. 1.8B, 7B, 14B, and 72B represent the model sizes. We computed the average accuracy across five domains for the model.}
\label{ZPT-results}
\end{table}

\subsection{Comparing Different Prompt Selection Strategies}
\label{Comparing Different Prompt Selection Strategies}

Experimental results are presented in Table \ref{tab-main-results-1}. We find that the prompt guidance constructed using our approach yields the highest translation quality across LLMs of different sizes, except for Qwen-1.8B-Chat. We attribute this to the fact that small models often struggle to extract meaningful context.

Additionally, in the same few-shot experimental setup, the translation quality of selecting demonstrations based on sentence embedding similarity surpasses that of the random approach. This is also corroborated in \cite{DBLP:conf/acl/AgrawalZLZG23}.
Additionally, we observe experimental results consistent with \cite{Radford2019LanguageMA}, indicating that the quality of generative outputs guided in a ``Zero-shot'' method is generally superior to those generated through ``Random'' selection or ``Similarity-based'' selection in a few-shot method.

We observe that ``Precedent'' serves as a strong baseline and exhibits a similar performance trend to our method: when the model size is too small, it struggles to produce high-quality translations. 
As noted by \citet{DBLP:conf/acl/AgrawalZLZG23}, the quality of demonstrations in prompt is crucial for translation performance.
Therefore, we attribute this to the cumulative effect of low-quality translations produced by small models, which leads to progressively worse translations.

Experiment results suggest that as the model size increases, our method continues to effectively guide LLMs in generating high-quality translations.
This also demonstrates the universality of our approach.

\begin{table*}[ht]
\centering
\begin{tabular}{lcccccccc}
\hline

\multicolumn{1}{l|}{\multirow{2}{*}{methods}}  & \multicolumn{2}{c}{Qwen-1.8B-Chat} & \multicolumn{2}{c}{Qwen-7B-Chat} & \multicolumn{2}{c}{Qwen-14B-Chat} & \multicolumn{2}{c}{Qwen-72B-Chat} \\
\cline{2-9}
\multicolumn{1}{l|}{} & test01 & test02 & test01 & test02 & test01 & test02 & test01 & test02 \\
 \hline
\multicolumn{1}{l|}{Zero-shot} & 13.77 & 8.93 & 19.33 & 14.09 & 21.43 & 15.52 & 22.96 & 16.61  \\
\multicolumn{1}{l|}{Random}    & 12.90 & 8.24 & 18.82 & 14.21 & 21.03 & 15.46 & 21.94 & 16.68  \\
\multicolumn{1}{l|}{Similar}   & 12.48 & 7.96 & 19.48 & 14.13 & 21.02 & 15.24 & 21.67 & 17.55  \\
\multicolumn{1}{l|}{Ours}       & \textbf{14.81} & \textbf{9.95} & \textbf{19.65} & \textbf{14.73} & \textbf{21.56} & \textbf{15.87} & \textbf{23.10} & \textbf{17.85} \\
\hline
\end{tabular}
\caption{Document-level BLEU scores on the WMT literary translation dataset, which includes two test sets, namely ``test01'' and ``test02''. ``test01'' and the datastores are within the same domain, while ``test02'' is in a different domain from the datastores.}
\label{LT-result}
\end{table*}

\subsection{Comparing with Traditional NMT Models}
We compared our method against traditional NMT models. As shown in Table \ref{encoder-decoder}, the results show that when the model size is large enough, the translation abilities of LLMs comprehensively exceed those of dedicated translation models.
This is observed except in the en-de language pair, where we attribute the discrepancy to the insufficient understanding of the German language by the Baichuan and Qwen models, primarily trained on Chinese and English corpora.
Specifically, in the en-zh translation direction where Baichuan and Qwen excel, the translation results of LLMs significantly surpass those of specialized models.

Moreover, our experiments reveal interesting insights into the impact of model size on translation performance. As expected, larger LLMs, such as Qwen-72B-Chat, consistently outperforms smaller counterparts across various language pairs and evaluation metrics. This trend demonstrates the importance of model scale in harnessing the full potential of LLMs for translation tasks.

However, the most striking finding is the competitive performance of small LLMs, particularly those with parameters in the 7B to 14B range. Despite their smaller size compared to their larger counterparts, these models exhibit translation capabilities on par with or even outperforming dedicated translation models. This suggests that with an appropriate prompt strategy, small LLMs can offer a cost-effective solution for translation tasks.

Our method has improved document translation performance on different LLMs, further verifying the robustness of our method. This underscores the generality and efficiency of our approach in leveraging LLMs for translation tasks across diverse domains and languages.

\subsection{Analyzing the Effectiveness on ZPT}

The increasing interest among researchers in ZPT can be attributed to its increasing complexity and significance within the field of language translation. ZPT involves the intricate task of proficiently addressing the omission of zero pronouns in the source language text when it undergoes translation into the target language. The primary aim is to ensure the meticulous restoration or expression of these omitted pronouns in the translated output, thereby upholding linguistic accuracy, fluency, and contextual consistency.

To assess the effectiveness of our proposed method in the context of the ZPT task, a series of experiments were conducted on the comprehensive GuoFeng dataset. Experimental results are displayed in Table \ref{ZPT-results}. 
Remarkably, across models of varying sizes, a discernible enhancement in ZPT accuracy is consistently observed when compared to the baseline performance. 
It is noteworthy, however, that the relationship between the model size and the accuracy of ZPT is not characterized by continuous improvement. 
This phenomenon can be ascribed to the realization that a high overall translation quality of the model does not necessarily translate into a proportionate increase in accuracy specifically related to ZPT.
Additionally, we found that the ``Precedent'' method outperforms similarity-based prompt selection strategies, indicating the importance of contextual information in the ZPT task. Our approach, which considers dynamic contextual information, effectively captures relevant context while filtering out irrelevant information. As a result, our method demonstrates superior performance in the ZPT task.

Hence, it becomes evident that a mere augmentation of the model size does not suffice for the improvement of ZPT accuracy. Consequently, the exploration and adoption of more effective methods become imperative in the pursuit of advancing the state-of-the-art in this domain.

\begin{figure*}[h]
\begin{center}
\includegraphics[scale=0.25]{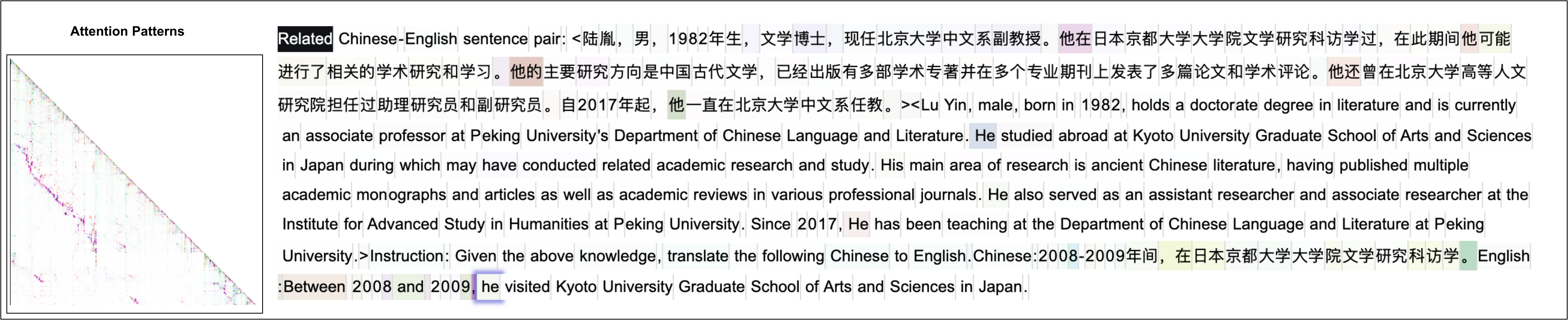} 
\caption{Average attention scores of the final layer of the Qwen-14B-Chat model. The darker the color, the higher the attention score assigned to that token. It is noteworthy that the model attends to ``他'' (meaning ``he'' in English) from the examples we extract when generating ``he''. Our examples successfully help the model in translating the zero pronoun ``他'' (meaning ``he'' in English).}
\label{attention}
\end{center}
\end{figure*}

\subsection{Analyzing the Effectiveness on Literary Translation}

The significant challenges for translating literary works include entity consistency, anaphora resolution, and lexical choice \cite{matusov-2019-challenges}. 
Literary works exhibit longer contextual spans compared to texts in other domains, such as news articles. Therefore, LLMs must possess the ability to model distant contexts in order to learn translation consistency and appropriate lexical choices.
Our approach, by focusing on the context relevant to the current sentence and subsequently summarizing the context, is able to encapsulate more contextual information within a limited-length prompt.
To validate whether our method can guide the model to generate more consistent translations, we conducted experiments on the WMT23 Literary Translation dataset.

The experimental results are presented in Table \ref{LT-result}. 
We observe consistent improvements of our approach over the baseline method, both on in-domain and out-of-domain test sets.
Consistent with the above experimental results, the strategy of simply using sentence embedding similarity to select demonstrations often performs even worse than random selection. This implies that a straightforward prompt selection strategy based on similarity is not suitable for challenging tasks like literary translation.
However, our method is able to effectively extract contextual information, thereby guiding LLMs to generate higher-quality translations, specifically achieving better entity consistency.

\begin{table}[!ht]
\centering
\begin{tabular}{lcc}
\hline
methods   & \multicolumn{1}{c}{de-en} & \multicolumn{1}{c}{zh-en} \\
\hline
\textit{w/o DCW}    & 35.84          & 26.03                   \\
\textit{max $\Rightarrow$ avg}    & 36.05          & 26.38                   \\
Ours       & \textbf{36.57} & \textbf{26.77}  \\
\hline
\end{tabular}
\caption{Ablation study on the de-en and zh-en WMT22 newstest dataset.}
\label{ablation-study}
\end{table}

\section{Analysis}
In this section, we conducted experiments to further analyze the effectiveness of our approach.

\subsection{Ablation Study}
To verify the effectiveness of various factors on our method, we further compared our method with the following variants and present the results in Table \ref{ablation-study}:

(1) \textit{w/o DCW}. In this variant, we directly used a Fixed Context Window to select the context. Specifically, we utilized the preceding and subsequent two sentences of the current sentence as the context. As reported in Table \ref{ablation-study}, this variant dramatically decreases performance on two language pair test sets. 
It reveals the significance of dynamically selecting context through attention scores.

(2) \textit{max $\Rightarrow$  avg}. In this variant, we replaced the max function with the average function in Eq.\;(\ref{token_sent_socre}). From Table \ref{ablation-study}, we can observe that this variant performs worse than our method. It demonstrates the effectiveness of our approach in computing the sentence-level attention score.

\subsection{Visualization}

In order to explain why our approach successfully translates omitted zero pronouns in Chinese, we extracted the attention weights from the final layer of the Qwen-14B-Chat model and visualized it using appropriate tools.\footnote{https://github.com/alan-cooney/CircuitsVis} 
As illustrated in Figure \ref{attention}, we observe that when the model attempts to translate the zero pronoun ``He'', it exhibits greater attention to the subject in the context and the associated information about the subject.
The visualized results reflect the effectiveness of our approach, showcasing its ability to capture relevant contextual information and guide the model in generating accurate and coherent translations.
This indicates that our approach involves more than merely translating words at the surface level; rather, it entails translating based on a profound comprehension of sentence structure and context, thereby offering guidance and support for the model to generate more precise translations, thereby demonstrating the effectiveness of our method.

\begin{table*}[h!]
\centering
\begin{tabular}{p{1.6cm}p{13.4cm}}
    \hline
    Source    & 主要从事媒体大数据计算、网络多媒体与跨媒体智能等研究工作。  \\
    Reference & \textcolor{blue}{His} research fields primarily concentrate on Media Big Data Computing, Network Multimedia and Cross-Media Intelligence, etc.    \\
    \hline
    Zero-shot & \textcolor{red}{Our} research focuses on media big data computing, network multimedia, and cross-media intelligence.  \\
    Random    & \textcolor{red}{My} main research works involve media big data computation, network multimedia and cross-media intelligence. \\
    Similar   & \textcolor{red}{The} main research areas include media big data computing, network multimedia and cross-media intelligence.  \\
    Ours      & \textcolor{blue}{His} main research focuses on media big data computation, network multimedia, and cross-media intelligence. \\
    \hline
  \end{tabular}
  \caption{A Case Study on the translation of Qwen-14B-Chat with various prompt selection strategies. \textcolor{red}{Red} text denotes incorrect translations of Zero Pronoun while \textcolor{blue}{blue} text indicates accurate translations.}
  \label{case-study}
\end{table*}

\subsection{Case Study}
In order to assess the precision of ZPT across various methods in the ZPT task, a comprehensive case study was undertaken. As illustrated in Table \ref{case-study}, it is observed that the ``Zero-shot'', ``Random'', and ``Similar'' methods exhibit inaccuracies in translating omitted zero pronouns in Chinese. Conversely, our approach demonstrates proficiency by accurately translating the omitted zero pronouns as ``His''. 
This success can be attributed to the method's adeptness at extracting relevant contextual information, enabling the model to deduce the omitted zero pronouns effectively.

In summary, our method stands out in its capacity to extract valuable contextual information, guiding the model towards generating translations that are not only more accurate but also more cohesive and coherent. Further examples and elaborate experimental results are available in Appendix \ref{Case Study and Visualization Analysis} for reference.

\section{Conclusion}

In this paper, we have presented a novel approach that guides the model to generate more accurate, cohesive, and coherent document-level translations via in-context learning. 
By selecting relevant contexts using our carefully designed method, we can extract topic information from the context summaries. These summaries serve as a guide for selecting the most beneficial examples from the datastore to enhance the in-context learning procedure of LLMs, hence mitigating the length limitation issue of in-context learning for DOCMT.
Experimental results on various benchmarks demonstrate the effectiveness of our approach, particularly in ZPT and literary translation tasks, where our method outperforms the baseline significantly.

\section*{Limitations}

By selecting better contexts and demonstrations, we have effectively increased the ZPT accuracy and BLEU score of DOCMT with LLMs via in-context learning. We have validated the effectiveness of our method using ZPT accuracy and BLEU score. However, beyond the general BLEU and Blonde metric, DOCMT encompasses challenges such as coreference consistency. We also need to make improvements to address these challenges. Moreover, we should also investigate the performance on more language pairs, which we consider as future work.

Yet another limitation in our work is the increased computational demand. Since our approach necessitates the computation of dynamic context and summarization of context, the computational load is slightly higher compared to the baseline. We intend to explore more efficient document-level machine translation methods in future work.

\section*{Acknowledgements}
The present research was supported by the National Key Research and Development Program of China (Grant No. 2023YFE0116400). We would like to thank the anonymous reviewers for their insightful comments.

% \bibliography{custom}

\begin{table*}[!ht]
\centering
\begin{tabular}{p{1.6cm}p{13.4cm}}
    \hline
    source    & 手动还是自动，确定离合掌握好了？  \\
    reference & Is your automatic or manual? Are \textcolor{blue}{you} sure that you deal with your clutch well?    \\
    \hline
    zero-shot & Should I use manual or automatic, and have \textcolor{red}{I} got a good grip on it?  \\
    random    & Should we use manual or automatic methods to solve the problem? Have \textcolor{red}{we} grasped the situation well? \\
    similar   & Should I  use manual or automatic control? Have \textcolor{red}{I} got a good grip on the distance and angle?  \\
    ours      & Manual or automatic, have \textcolor{blue}{you} determined the gear shift handle is in the correct position? \\
    \hline
  \end{tabular}
  \caption{A Case Study on the translation of Qwen-14B-Chat with various prompt selection strategies. \textcolor{red}{Red} text denotes incorrect translations of Zero Pronoun while \textcolor{blue}{blue} text indicates accurate translations.}
  \label{appendix-case-study}
\end{table*}

\begin{figure*}[!ht]
\begin{center}
\includegraphics[scale=0.23]{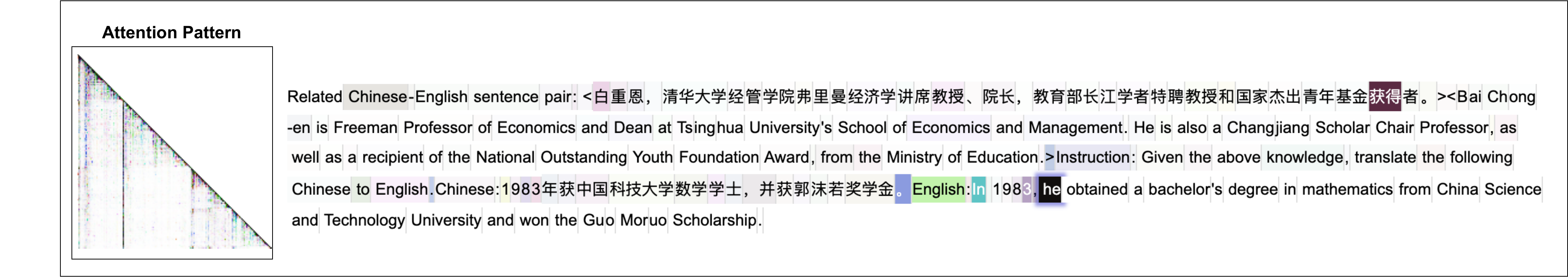} 
\caption{Average attention scores of the final layer of the Qwen-14B-Chat model. The darker the color, the higher the attention score assigned to that token.}
\label{attention-appendix}
\end{center}
\end{figure*}

\newpage

\appendix
\section{Additional Experimental Results}
\label{Experimental results}

\begin{table}[ht]

\begin{tabular}{lllll}
\hline
\multirow{2}{*}{methods} & \multicolumn{2}{l}{Qwen-7B-Chat} & \multicolumn{2}{l}{Qwen-14B-Chat} \\
& de-en           & zh-en          & de-en           & zh-en           \\
\cline{1-5}
Zero-shot & 46.88           & 35.15          & 50.01           & 38.27           \\
Random    & 46.47           & 34.74          & 49.35           & 38.27           \\
BM25      & 45.94           & 34.63          & 49.37           & 38.00           \\
Similar   & 46.36           & 34.77          & 49.72           & 38.82           \\
Precedent & 46.13           & 35.04          & 49.50           & 38.43           \\
Ours      & \textbf{47.00}  & \textbf{35.20} & \textbf{50.23}  & \textbf{38.90}  \\
\hline
\end{tabular}
\caption{Blonde scores of Qwen chat LLMs, with different prompt strategies, on the WMT22 newstest dataset.}
\label{blonde-appendix}
\end{table}

In Section \ref{Comparing Different Prompt Selection Strategies}, we have employed only $d$-BLEU and ChrF2 to evaluate  translation quality. We use Blonde scores here to further measure translation quality.

Specifically, we evaluated the translation performance of Qwen-7B-Chat and Qwen-14B-Chat on the WMT22 newtest test set using the Blonde metric under different prompt strategies. Experimental results, shown in Table \ref{blonde-appendix}, indicate that our proposed method significantly outperforms the baselines, demonstrating its effectiveness.

\section{Case Study and Visualization Analysis}
\label{Case Study and Visualization Analysis}

In this section, we present more examples to further validate our approach with case study and attention visualization.

\subsection{Case Study}

As shown in Table \ref{appendix-case-study}, we provide another example where in the original Chinese text, the subject ``你'' (you) is omitted due to language convention. In English, it should be correctly restored and translated as ``you''. We find that the ``Zero-shot'', ``Random'', and ``Similar'' methods all incorrectly recover zero pronouns, whereas our method successfully recovers the correct zero pronoun ``you''. This demonstrates that our method outperforms better than baselines in ZPT.

\subsection{Visualization}
As illustrated in Figure \ref{attention-appendix}, we observe that when the model is tasked with translating ``He'', it predominantly focuses on the subject of the context and its associated information, encompassing entities such as ``白崇恩'' and the designation ``教授''. 
This experimental result illustrates how our approach guides LLMs to generate more accurate translations via in-context learning, thus proving the effectiveness of our method.

\end{CJK}
\end{document}